\documentclass{article}

\usepackage[T1]{fontenc}
\usepackage{graphicx}
\usepackage{url}
\usepackage{amsmath,amssymb,amsfonts}
\usepackage{algorithmic}
\usepackage{textcomp}
\usepackage{xcolor}
\usepackage{tikz}
\usetikzlibrary{shapes, arrows, backgrounds, intersections, decorations.pathreplacing, mindmap,trees, positioning, arrows.meta}
\usepackage{csvsimple}
\usepackage{pdflscape}
\usepackage{caption}
\usepackage{subcaption}

\usepackage{xypic}
\usepackage{hyperref}
\usepackage{cleveref}
\usepackage{xspace}
\newcommand{\wdn}{WDN\xspace}
\newcommand{\wdns}{WDNs\xspace}
\newcommand{\ie}{i.e.\xspace}
\newcommand{\eg}{e.g.\xspace}
\newcommand{\ml}{ML\xspace}

\newcommand{\Predbased}{Prediction-residual-based\xspace}
\newcommand{\Obsbased}{Observation-residual-based\xspace}

\usepackage{csquotes}
\usepackage{pifont}
\usepackage{pifont}
\usepackage{booktabs}
\usepackage{hyperref}
\usepackage{multirow}
\usepackage{multicol}
\usepackage{nicefrac}
\usepackage{varwidth}
\usepackage{pgfplotstable}
\pgfplotsset{compat=1.16}
\usepackage{tabularx}
\usepackage{adjustbox}
\makeatletter
\pgfplotsset{
    /pgfplots/table/omit header/.style={%
        /pgfplots/table/typeset cell/.append code={%
            \ifnum\c@pgfplotstable@rowindex=-1
                \pgfkeyslet{/pgfplots/table/@cell content}\pgfutil@empty%
            \fi
        }
    }
}
\makeatother

\usepackage[binary-units,group-separator={;}]{siunitx}

\newcolumntype{N}{S[table-format=1.2,round-mode=places,round-precision=2]}
\newcolumntype{V}{S[table-format=3.1,round-mode=places,round-precision=1]}
\newcolumntype{O}{S[table-format=1.3]}
\newcolumntype{Q}{@{\,}c@{\,}S[table-format=1.1,round-mode=places,round-precision=1]@{}c@{\ }}
\newcolumntype{W}{@{\,}c@{\,}S[table-format=3.1,round-mode=places,round-precision=1]@{}c}
\newcolumntype{X}{@{\,}c@{\,}S[table-format=2.1,round-mode=places,round-precision=1]@{}c}
\newcolumntype{R}{@{\,}c@{\,}S[table-format=1.2,round-mode=places,round-precision=2]@{}c@{\ }}

\newcolumntype{L}{S[table-format=4.1,round-mode=places,round-precision=1]}
\newcolumntype{K}{S[table-format=1.2]}
\newcolumntype{S}{S[table-format=1.1,round-mode=places,round-precision=1]}
\newcolumntype{M}{@{\,}c@{\,}S[table-format=3.1,round-mode=places,round-precision=1]@{}c}

\newcommand{\overviewtab}[3][]{
\setlength{\tabcolsep}{0.25em}
    \ifthenelse{\equal{#3}{}}{}
    {{\centering\tiny\textbf{#3}\\\medskip}}
    \pgfplotstabletypeset[
    col sep=semicolon,
    string type,
    columns/References/.style={string type,column type={@{\ }l@{\ }}, column name={ref}},
    columns/Networks/.style={string type,column type={@{\ }|m{3.3em}@{\ }}, column name={}},
    columns/historicdata/.style={string type,column type={@{\ }|>{\centering}m{3.7em}@{\ }}, column name={leak-free historic data}},
    columns/topology/.style={string type,column type={@{\ }>{\centering}m{3.7em}@{\ }}, column name={precise topology}},
    columns/demands/.style={string type,column type={@{\ }>{\centering}m{3.7em}@{\ }}, column name={real-time demand}},
    columns/prediction/.style={string type,column type={@{\ }|m{14em}@{\ }}, column name={predict ($h$)/process}},
    columns/detection/.style={string type,column type={@{\ }m{14em}@{\ }}, column name={detect}},
    columns/localization/.style={string type,column type={@{\ }|m{13em}@{\ }}, column name={}},
    columns/historical/.style={string type,column type={@{\ }c@{\ }}, column name={strategy}},
    columns={References, historical, historicdata, topology, demands,prediction, detection,localization, Networks},
    every head row/.style={after row=\midrule, before row={\multicolumn{2}{c|}{Method}&\multicolumn{3}{c|}{Data requirements}&\multicolumn{2}{c|}{Detection}&\multicolumn{1}{c|}{Localization}&\multicolumn{1}{c}{Data}\\}},
    ]{#2}
}

\newcommand{\networktab}[3][]{
    \ifthenelse{\equal{#3}{}}{}
    {{\centering\tiny\textbf{#3}\\\medskip}}
    \pgfplotstabletypeset[
    col sep=semicolon,
    string type,
    columns/Network/.style={string type,column type={@{\ }l@{\ }|}, column name={Network}},
    columns/ref/.style={string type,column type={@{\ }r@{\ }|}, column name={References}},
    columns/nodes/.style={string type,column type={@{\ }r@{\ }|}, column name={$\#$ nodes}},
    columns/pipes/.style={string type,column type={@{\ }r@{\ }}, column name={$\#$ pipes}},
    columns={Network, ref, nodes, pipes},
    every head row/.style={after row=\midrule},
    ]{#2}
}
\newcommand{\atmntab}[3][]{
    \ifthenelse{\equal{#3}{}}{}
    {{\centering\tiny\textbf{#3}\\\medskip}}
    \pgfplotstabletypeset[
    col sep=semicolon,
    string type,
    omit header,
    columns/type/.style={string type,column type={@{\ }l@{\ }|}, column name={}},
    columns/descr/.style={string type,column type={@{\ }m{20em}@{\ }}, column name={}},
    columns={type,descr},
    ]{#2}
}

\newcommand{\resultstab}[2][]{
    \pgfplotstabletypeset[
    outfile=tab-fault.tex,
    col sep=comma,
    string type,
    columns/us_delay/.style={column type=X,column name=},
    columns/lb_delay/.style={column type=W,column name=},
    columns/xu_delay/.style={column type=W,column name=},
    columns/error/.style={column type=Q,column name=},
    columns/us_dist_node/.style={column type=Q,column name=},
    columns/lb_dist_node/.style={column type=Q,column name=},
    columns/us_delay_ave/.style={column type=V,column name=\multicolumn{3}{c}{\textbf{ours}}},
    columns/lb_delay_ave/.style={column type=V,column name=\multicolumn{3}{c}{\textbf{LB}}},
    columns/xu_delay_ave/.style={column type=V,column name=\multicolumn{3}{c}{\textbf{DEIM}}},
    columns/error_ave/.style={column type=N,column name=\multicolumn{3}{c}{\textbf{error}}},
    columns/us_dist_node_ave/.style={column type=S,column name=\multicolumn{3}{c}{\textbf{ours}}},
    columns/lb_dist_node_ave/.style={column type=S,column name=\multicolumn{3}{c}{\textbf{LB}}},
    columns/leak_dia/.style={column type=l,column name={\textbf{leak dia}}},
    columns/us_FP/.style={column type=r,column name={\textbf{ours}}},
    columns/lb_FP/.style={column type=r,column name={\textbf{LB}}},
    columns/xu_FP/.style={column type=S,column name=\multicolumn{3}{c}{\textbf{DEIM}}},
    columns/xu_FP_all/.style={column type=R,column name=},
    columns/us_TP/.style={column type=r,column name={\textbf{ours}}},
    columns/lb_TP/.style={column type=r,column name={\textbf{LB}}},
    columns/xu_TP/.style={column type=r,column name={\textbf{DEIM}}},
    columns={leak_dia,us_TP,lb_TP,xu_TP,us_delay_ave,us_delay,lb_delay_ave,lb_delay,xu_delay_ave,xu_delay,us_FP,lb_FP,xu_FP,xu_FP_all,us_dist_node_ave,us_dist_node,lb_dist_node_ave,lb_dist_node},%
    every head row/.style={after row=\midrule,before row={&\multicolumn{3}{c}{\textbf{$\#$detects/20 scenarios}}&\multicolumn{12}{c}{\textbf{detection delay}}&\multicolumn{6}{c}{\textbf{false alarms}}&\multicolumn{7}{c}{\textbf{topological distance}}\\[0.3em]}},
    #1,
    create on use/us_delay/.style={
      create col/assign/.code={%
        \ifnum\pdfstrcmp{\thisrow{us_delay_ave}}{}=0
        \edef\entry{--&&}
        \else
        \edef\entry{$\pm$&\thisrow{us_delay_std}&}
        \fi
        \pgfkeyslet{/pgfplots/table/create col/next content}\entry
      }
    },
    create on use/lb_delay/.style={
      create col/assign/.code={%
        \ifnum\pdfstrcmp{\thisrow{lb_delay_ave}}{}=0
        \edef\entry{--&&}
        \else
        \edef\entry{$\pm$&\thisrow{lb_delay_std}&}
        \fi
        \pgfkeyslet{/pgfplots/table/create col/next content}\entry
      }
    },
    create on use/xu_delay/.style={
      create col/assign/.code={%
        \ifnum\pdfstrcmp{\thisrow{xu_delay_ave}}{}=0
        \edef\entry{--&&}
        \else
        \edef\entry{$\pm$&\thisrow{xu_delay_std}&}
        \fi
        \pgfkeyslet{/pgfplots/table/create col/next content}\entry
      }
    },
    create on use/us_dist_node/.style={
      create col/assign/.code={%
        \ifnum\pdfstrcmp{\thisrow{us_dist_node_ave}}{}=0
        \edef\entry{--&&}
        \else
        \edef\entry{$\pm$&\thisrow{us_dist_node_std}&}
        \fi
        \pgfkeyslet{/pgfplots/table/create col/next content}\entry
      }
    },
    create on use/lb_dist_node/.style={
      create col/assign/.code={%
        \ifnum\pdfstrcmp{\thisrow{lb_dist_node_ave}}{}=0
        \edef\entry{--&&}
        \else
        \edef\entry{$\pm$&\thisrow{lb_dist_node_std}&}
        \fi
        \pgfkeyslet{/pgfplots/table/create col/next content}\entry
      }
    },
    create on use/xu_FP_all/.style={
      create col/assign/.code={%
        \ifnum\pdfstrcmp{\thisrow{xu_FP}}{}=0
        \edef\entry{--&&}
        \else
        \edef\entry{$\pm$&\thisrow{xu_FP_std}&}
        \fi
        \pgfkeyslet{/pgfplots/table/create col/next content}\entry
      }
    },
    ]{#2}
}

\begin{document}

\title{Challenges, Methods, Data -- a Survey of Machine Learning in Water Distribution Networks\footnote{This preprint has
not undergone any post-submission improvements or corrections. The Version of Record of this contribution is published in Artificial Neural Networks and Machine Learning – ICANN 2024,
and is available online at \url{https://doi.org/10.1007/978-3-031-72356-8_11}}}

\author{
Valerie Vaquet,
Fabian Hinder,
André Artelt, \\
Inaam Ashraf, 
Janine Strotherm,
Jonas Vaquet, \\
Johannes Brinkrolf,
Barbara Hammer \\\;\\ \small Bielefeld University, Bielefeld, Germany}

\maketitle       
\begin{abstract}
Research on methods for planning and controlling water distribution networks gains increasing relevance as the availability of drinking water will decrease as a consequence of climate change. So far, the majority of approaches is based on hydraulics and engineering expertise. However, with the increasing availability of sensors, machine learning techniques constitute a promising tool. This work presents the main tasks in water distribution networks, discusses how they relate to machine learning and analyses how the particularities of the domain pose challenges to and can be leveraged by machine learning approaches. Besides, it provides a technical toolkit by presenting evaluation benchmarks and a structured survey of the exemplary task of leakage detection and localization.

\textbf{keywords: }{Water Distribution Networks $\cdot$ Survey $\cdot$ Concept Drift.}
\end{abstract}
\section{Introduction}
High levels of threat in water security concern almost 80\% of the world's population~\cite{vorosmarty2010global}. Recent studies show that this effect will aggravate as water resources become more scarce due to climate change~\cite{rodell2018emerging}.
Having well-working water distribution networks (WDNs) in place plays a crucial role in using the limited resources most efficiently and ensuring the quality of the available drinking water. 
As reliable and clean drinking water is essential to the health and well-being of the population, similar to electrical grids and transportation systems, \wdns are considered to be part of the critical infrastructure \cite{eliades_fault_2010}.
As a consequence, approaches using artificial intelligence (AI) need to obey specific regulations on safety, robustness, and human agency as specified in the European AI-ACT \cite{EU_proposal_AI}.

Since planning and controlling these systems is a non-trivial task, \wdns are an active research area for experts from hydro-informatics and control theory \cite{mala-jetmarova_lost_2018,eliades_fault_2010,hu_survey_2018} with multiple scientific challenges being hosted, e.g. \cite{marchi_battle_2014,taormina_battle_2018,vrachimis_battle_2022}. While systems are traditionally planned, maintained, and controlled using engineering expertise -- e.g. by modeling and control based on knowledge about the hydraulic and chemical processes in the system -- with increasing availability of (mobile) sensor devices, machine learning (ML) techniques gain relevance as they can have a profound impact in this area \cite{hu_review_2021}. Facing upcoming challenges that will expand in the future \cite{rodell2018emerging} contributions by ML experts have the potential to accelerate research efforts in this domain. Besides, as the domain has some specific properties causing challenges but also chances for ML approaches, considering \wdns as a benchmark scenario to evaluate algorithmic contributions is promising.

The goal of this work is to provide a formalization and structured survey of challenges in the \wdn domain which can be targeted as \ml tasks, and references to possible benchmark data. To the best of our knowledge so far no such survey exists. In contrast to \cite{denakpo_machine_2024}, which performs a bibliometric analysis on the body of the literature, as our main contributions, we outline and structure the characteristics of the domain with a focus on ML applications, propose how to connect the main tasks in WDNs to classical ML tasks, and specify the main challenges and chances induced by the characteristics of the domain.
Besides, focusing on the exemplary task of leakage detection and localization as an exemplary assignment
    we structure the current state-of-the-art methods from the perspective of \emph{drift detection and analysis} and examine the strengths and limitations of different methodological families.
    Additionally, we provide an overview of benchmarks and suitable evaluation strategies for this task.
This survey equips ML researchers with a toolkit to work in the domain of WDNs, with our more detailed analysis of leakage detection and localization serving as a blueprint for the other presented tasks.

This paper is structured as follows: Section~\ref{sec:domain} discussed the characteristics of the water domain while Section \ref{sec:tasks} provides an overview of the main tasks in the domain, alongside the connected challenges and opportunities. Then, focusing on leakage detection and localization, we present evaluation benchmarks (Section \ref{sec:data}), existing solution schemes, and categorize the latter (Section \ref{sec:approaches}).

\section{Particularities of the Domain\label{sec:domain}}

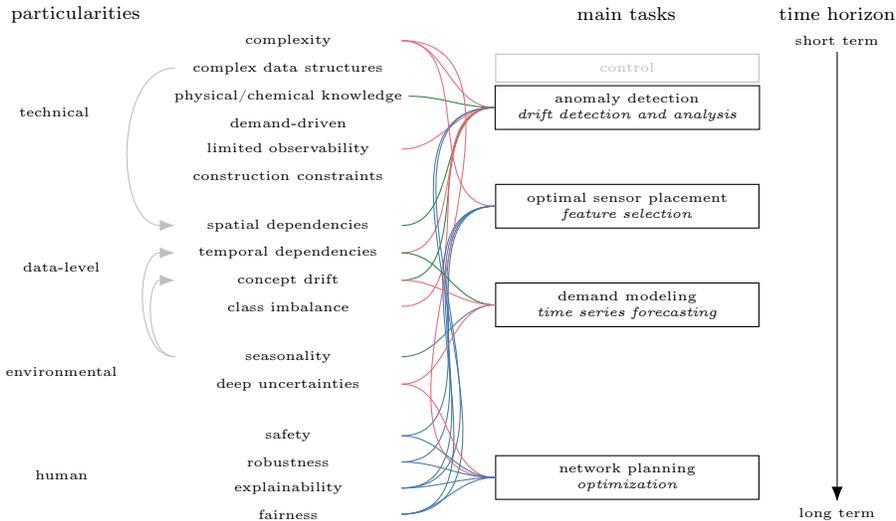
\begin{figure}[t]
    \definecolor{mygreen}{RGB}{34,136,51}
    \definecolor{myred}{RGB}{238,102,119}
    \definecolor{myblue}{RGB}{68,119,170}
    \begin{center}
    \tiny
    \hspace{-4em}
    \begin{tikzpicture}[node distance=0.15em, minimum width=3cm]
        \node at (0,0) [] (complexity){complexity};
        \node[below = of complexity] [] (graph){complex data structures};
        \node[below = of graph] [] (domain){physical/chemical knowledge};
        \node[below = of domain] [] (demand){demand-driven};
        \node[below = of demand] [] (obs){limited observability};
        \node[below = of obs] [] (construction){construction constraints};
        \node[below left = -0.5em and 0em of domain] [] (technical){technical};
        
        \node[below = 1.5em of construction] [] (spatial){spatial dependencies};
        \node[below = of spatial] [] (temporal){temporal dependencies};
        \node[below = of temporal] [] (drift){concept drift};
        \node[below = of drift] [] (imb){class imbalance};
        \node[below left = -0.5em and 0em  of temporal] [] (data){data-level};

        \node[below = 1.5em of imb] [] (seasonality){seasonality};
        \node[below = of seasonality] [] (uncertainty){deep uncertainties};
        \node[below left = -0.5em and 0em  of seasonality] [] (env){environmental};

        \node[below = 1.5em of uncertainty] [] (safety){safety};
        \node[below = of safety] [] (robustness){robustness};
        \node[below = of robustness] [] (explainability){explainability};
        \node[below = of explainability] [] (fairness){fairness};
        \node[below left = -0.5em and 0em  of robustness] [] (human){human};

        \draw[-{Latex[length=2mm]}] (graph) edge [out=180,in=180, lightgray] (spatial);
        \draw[-{Latex[length=2mm]}] (seasonality) edge [out=180,in=180, lightgray] (temporal);
        \draw[-{Latex[length=2mm]}] (seasonality) edge [out=180,in=180, lightgray] (drift);

        \node[right = 20em of complexity, minimum width=2cm](short){short term};
        \node[right = 20em of fairness, minimum width=2cm](long){long term};
        \draw [-{Latex[length=2mm]}] (short.south) -- (long.north);
        \node[below left =  of short, rectangle, draw=lightgray, minimum width=3.5cm] (control)
        {\begin{tabular}{c}
             \textcolor{lightgray}{control}
        \end{tabular}};
        \node[below = of control, rectangle, draw=black, minimum width=3.5cm] (anomalydetection)
        {\begin{tabular}{c}
             anomaly detection\\
             \emph{drift detection and analysis} 
        \end{tabular}};
        \node[below = 3em of anomalydetection, rectangle, draw=black, minimum width=3.5cm] (sensorplacement){\begin{tabular}{c}
             optimal sensor placement\\
             \emph{feature selection} 
        \end{tabular}};
        \node[below = 3em of sensorplacement, rectangle, draw=black, minimum width=3.5cm](demand)
        {\begin{tabular}{c}
             demand modeling\\
             \emph{time series forecasting} 
        \end{tabular}};
        \node[above left = of long, rectangle, draw=black, minimum width=3.5cm] (network)
        {\begin{tabular}{c}
             network planning\\
             \emph{optimization} 
        \end{tabular}};

        \node[above = of short, minimum width=2cm](time){\scriptsize time horizon};
        \node[left = of time, minimum width=3.5cm](tasks){\scriptsize main tasks};
        \node[left =17em of tasks]{\scriptsize particularities};

        \draw[-, color=mygreen] (domain) edge [out=0,in=180] (anomalydetection);
        \draw[-, color=mygreen] (drift) edge [out=0,in=180] (anomalydetection);
        \draw[-, color=mygreen] (spatial) edge [out=0,in=180] (anomalydetection);
        \draw[-, color=myred] (temporal) edge [out=0,in=180] (anomalydetection);
        \draw[-, color=myred] (complexity) edge [out=0,in=180] (anomalydetection);
        \draw[-, color=myred] (obs) edge [out=0,in=180] (anomalydetection);
        \draw[-, color=myred] (imb) edge [out=0,in=180] (anomalydetection);
        \draw[-, color=myblue] (fairness) edge [out=0,in=180] (anomalydetection);
        \draw[-, color=myblue] (explainability) edge [out=0,in=180] (anomalydetection);

        \draw[-, color=myred] (complexity) edge [out=0,in=180] (sensorplacement);
        \draw[-, color=myblue] (safety) edge [out=0,in=180] (sensorplacement);
        \draw[-, color=myblue] (robustness) edge [out=0,in=180] (sensorplacement);
        \draw[-, color=myblue] (explainability) edge [out=0,in=180] (sensorplacement);
        \draw[-, color=myblue] (fairness) edge [out=0,in=180] (sensorplacement);
        
        \draw[-, color=mygreen] (temporal) edge [out=0,in=180] (demand);
        \draw[-, color=myred] (drift) edge [out=0,in=180] (demand);
        \draw[-, color=myblue] (seasonality) edge [out=0,in=180] (demand);
        \draw[-, color=myred] (uncertainty) edge [out=0,in=180] (demand);

        \draw[-, color=myred] (uncertainty) edge [out=0,in=180] (network);
        \draw[-, color=myred] (complexity) edge [out=0,in=180] (network);
        \draw[-, color=myblue] (fairness) edge [out=0,in=180] (network);
        \draw[-, color=myblue] (explainability) edge [out=0,in=180] (network);
        \draw[-, color=myblue] (robustness) edge [out=0,in=180] (network);
        \draw[-, color=myblue] (safety) edge [out=0,in=180] (network);
        
    \end{tikzpicture}
    \caption{Overview of the particularities of and key tasks in WDNs (in the boxes) and their dependencies.
    Red connections mark key challenges, green potentials, and blue additional constraints.\label{fig:overview} }
    \end{center}
\end{figure}

When designing ML approaches for the domain of \wdns, one needs to account for several particularities, which can pose challenges and offer potential for possible solutions. As \wdns are part of the critical infrastructure, next to technical particularities, one needs to account for environmental and human factors. These give rise to data-level aspects which need to be accounted for as well. We briefly discuss these key points which are summarized in the left part of Fig. \ref{fig:overview}.

\subsection{Technical Aspects}
The \emph{complexity of problems} in \wdns is a key challenge. WDNs are very complex systems with many different components (e.g. valves, pumps, and reservoirs) which can change the direction of the water flow and the system's dynamics. This yields mixed integer/real-valued problems which are paired with \emph{complex data structures} such as complex dynamic graphs \cite{mala-jetmarova_lost_2018}.
While the hydraulic dynamics pose challenges, they also offer opportunities: As the dynamics in the networks underly the \emph{laws of physics and chemistry}, models might be improved by considering physics-informed \ml and incorporating domain-specific engineering expertise.
However, \wdns are generally \emph{demand-driven systems}. Thus, a system is highly influenced by the behavior of customers resulting in daily and weekly patterns that inflict themselves on the system. Another obstacle is the \emph{limited observability} of water networks. Due to the associated cost and installation efforts, \wdns are only starting to be scarcely equipped with sensor technology~\cite{eggimann_potential_2017}.

\subsection{Environmental Aspects}
Next to daily and weekly patterns, the water demand is also influenced by environmental effects, e.g. \emph{seasonalities}, and short-term factors, e.g. the weather conditions or vacation times \cite{donkor_urban_2014}. On a longer time horizon, one needs to account for \emph{deep uncertainties} \cite{gass_deep_2013}: As we cannot describe environmental and societal developments in the future, we often have to assume a range of possible future scenarios when working on long-term solutions.

\subsection{Data-Level Aspects}
The discussed technical and environmental aspects yield some data-level aspects. While the complex graph structure results in \emph{spatial dependencies}, seasonal effects and the influence of demands result in \emph{temporal dependencies}. Besides, seasonalities and rare events can result in distributional changes, also referred to as \emph{concept drift} \cite{gama2014survey}. Finally, \emph{class imbalance} in the measurements has to be expected, as usually much fewer measurements of anomalous behavior or rare events such as leakages or extreme weather are collected.
This can pose a problem for ML-based approaches.
 
\subsection{Human Factor\label{sec:parti-society}}
In addition to the more technical aspects, some additional constraints are imposed on \wdns as there are additional societal requirements on critical infrastructure. Since physical destruction or poisoning pose high risks as contaminated water can spread in large areas and may threaten the health of many people \cite{hu_survey_2018} and \wdns are vulnerable to attacks due to their size, paying attention to safety when developing methods is crucial. Thus, ensuring the \emph{safety} and \emph{robustness} of the developed approaches is a key criterium. 

Besides, according to the European AI-ACT~\cite{EU_proposal_AI}, \emph{explainability}~\cite{molnar2020interpretable} and \emph{fairness}~\cite{castelnovo2022clarification} need to be ensured. Providing explainable AI models not only increases the acceptance of the general public but also facilitates the possibility of blending domain knowledge provided by human experts with the models. Further, explainability enables verification in case models with unclear black-box generalization ability are used.
As it is crucial to provide reliable services to different locations with varying demographics throughout a city, considering the fairness of \ml models is essential when developing new methods.

\section{Tasks in Water Distribution Networks\label{sec:tasks}}

The main tasks in \wdns can be categorized according to their time horizon ranging from \emph{real-time control} to the \emph{long-term planning} of the network layout as visualized in the right part of Fig.~\ref{fig:overview}. In this section, we will introduce them, propose how they can be connected to classical ML assignments, and discuss which particularities need to be accounted for when designing approaches. Note that we will not cover the task of control due to space constraints.

\subsection{Anomaly Detection}
Next to the actual operation of \wdns, \emph{real-time control} systems aim at supervising the network and resolving anomalous behavior. Anomalies generally can be categorized into hydraulic anomalies, e.g. leakages, and water quality anomalies, e.g. contamination~\cite{eliades_contamination_2023}. Both might occur by accident or be caused intentionally. Usually, the implementation of monitoring systems is split into the following subtasks~\cite{eliades_fault_2010}:
    First, anomalies need to be detected, \ie an alarm should be raised whenever there is atypical behavior in the system.
    In the second step, more information about the anomalous event is collected. The goal is to identify the type of anomaly (\eg sensor fault, leakage, pipe burst, fire hydrant usage), localize its position in the network, and determine the exact starting time of the anomaly.
    Finally, some kind of accommodation is required. This can range from physical actions, \eg repairing a leaking pipe, to software solutions, \eg imputing missing data in case of failing sensors \cite{osman_survey_2018}.

\ml techniques are increasingly used for anomaly detection as they require less knowledge than traditional engineering approaches.
In contrast, localizing the anomalies in the network frequently relies mostly on hydraulic approaches.
 
From a \ml perspective, modeling these tasks as \emph{drift detection and analysis} is suitable~\cite{vaquet2023localization}. While this modeling exploits the fact that anomalies inflict themselves as concept drift, solutions need to account for the complexity, and limited observability of \wdns. Besides, temporal dependencies and class imbalances concerning anomalies pose challenges. In converse, it might be beneficial to leverage domain knowledge and spatial dependencies when designing methodologies. Fairness and explainability requirements should be met. We discuss leakage detection and localization as an exemplary assignment in Section~\ref{sec:approaches}.

Imputation for missing or faulty sensor measurements has mostly been done by standard imputation techniques \cite{osman_survey_2018}. Recently using virtual sensors has been proposed, e.g., by simple linear models \cite{DBLP:conf/icann/VaquetABH22} or graph convolutional networks \cite{ashraf2022gcn_wds}.

\subsection{Optimal Sensor Placement}
To implement the discussed tasks of anomaly detection and control, the installation of adequate sensor technology is required. Optimal sensor placement focuses on how to find an appropriate placement of sensors across the network.

Finding an optimal sensor placement is a difficult task due to the complexity of the network. 
The majority of work in this field focuses on heuristics and optimization methods~\cite{hu_survey_2018}. Many approaches rely on having a simulation of the network available to evaluate the suitablility of different sensor placements for downstream tasks like anomaly detection. Besides, most works assume that the sensors have perfect measurement capabilities, e.g., that quality sensors pick up even the smallest concentration of a given substance \cite{hu_survey_2018}. Recently, some initial works incorporated \ml methods into the optimization process~\cite{candelieri_lost_2022}.

Assuming a simulation is available, from an ML perspective, this task can be formalized as a \emph{feature selection} problem, where all possible sensor types and locations are features to select from. However, when performing feature selection in \wdns, this should ideally be done with multiple downstream tasks in mind as finding one suitable sensor placement reduces cost and construction efforts. For example, it would be desirable to find a placement putting hydraulic and quality sensors in the same spots as each installation requires construction work to access the pipes. Additionally, due to the nature of the domain of \wdns, this sensor placement needs to enhance safety and be robust, explainable, and fair.

\subsection{Demand Modeling}
Modeling demands on short time horizons is important to inform real-time operations while on long time horizons, it is key to planning \wdns.
Modeling this task, we can benefit from the property of temporal dependencies in the data: We can assume time series data, making methods from \emph{time-series forecasting} available. However, one needs to account for seasonality and concept drift. Considering long time horizons deep uncertainties pose additional challenges.

There are some approaches relying on \ml with the majority being neural network models for time series prediction or regression~\cite{alvisi2017}. \ml-based methods have the advantage of making fewer assumptions than conventional models~\cite{donkor_urban_2014}. Focusing on short-term modeling, applying adaptive methods is promising as less data is required initially and trends can be captured more successfully. Here, \ml techniques are not widely explored yet~\cite{pacchin_comparison_2019}.

\subsection{Long-Term Planning}

\emph{Long-term planning} of \wdns aims at finding an optimal network topology given multiple constraints while accounting for deep uncertainties, e.g., the expansion of cities, and external factors, e.g., climate change, are not clearly predictable.
Planning is a very complex \emph{optimization} problem. In addition to the underlying complexity, requirements regarding safety and robustness, and deep uncertainties, one needs to consider additional constraints on fairness and explainability. 

Considerable research has been conducted on planning under these uncertainties by so-called staged and flexible design~\cite{mala-jetmarova_lost_2018,tsiami_review_2022}. While intelligent optimization technologies such as evolutionary strategies have extensively been used to target the task of planning as an optimization problem~\cite{mala-jetmarova_lost_2018}, \ml methods have only occasionally been integrated mostly for medium or short-term subtasks such as prioritizing pipelines for rehabilitation~\cite{elshaboury_prioritizing_2022}.

In the remainder of this paper, we will exemplarily discuss the tasks of leakage detection and localization as there exist already several \ml approaches due to available sensor data and benchmark scenarios for evaluation. The specific modeling for dealing with leakages can serve as a blueprint for other challenges, where training samples for \ml are not as readily available yet.

\section{Data and Evaluation for Anomaly Detection\label{sec:data}}

As mentioned before evaluation benchmarks for the tasks of leakage detection and localization are available. We will present available data benchmarks and commonly used evaluation schemes.
There are three potential options for evaluating leakage detection and localization algorithms concerning data sets. (i)~\emph{Real-world measurements} are difficult to collect and frequently not publicly available. Besides, usually, it is not possible to obtain reliable ground truth about the underlying anomalies, e.g., it is hard to figure out the precise starting point of a small leak. Further, data-related particularities as discussed in Section~\ref{sec:domain} like limited observability and class imbalance pose problems.
For this reason, research projects in this domain frequently rely on (ii)~simulated data. While there exists (ii-a)~\emph{a collection of ready-to-use simulated scenarios} which are available as part of LeakDB \cite{authors_vrachimis_leakdb_2018}, it might be favorable to (ii-b)~\emph{generate custom scenarios} as one can control the exact type, timing, and location of the anomalies and generate multiple similar scenarios to evaluate novel methodologies more thoroughly.

We will briefly describe how \wdns are modeled, as this is a necessary prerequisite for the data simulation process. Besides, we present both standard network layouts and the role of demand patterns when simulating data and cover evaluation strategies.

\textbf{Modeling WDNs} Water supply networks can be modeled as graphs consisting of nodes representing junctions and undirected edges representing pipes as the flow direction of the water is not pre-defined and can change over time due to different inputs and demands in the network. As usually real-world water supply systems are extended over time, it is possible that multiple pipes lie parallel to each other \cite{mala-jetmarova_lost_2018}. Hence, in this domain, we deal with multigraphs that might change over time. 
Frequently, water systems also contain some sensor technology that can be installed at the connections of the pipes or in basins (\eg level or pressure sensors) or in the pipes (\eg flow sensors). While it is also possible to measure the demand of the end-users by so-called smart meters, usually this is not done exhaustively due to the associated cost and privacy concerns \cite{smart-meters}.

\textbf{WDNs and simulation tools}
There are few popular openly available water networks. They range from very small systems like the Hanoi with 32 nodes and 34 pipes or the Anytown network with 20 pipes and 34 nodes to larger, more realistic ones like L-Town containing 785 nodes and 909 pipes. Some scenarios are available as part of LeakDB \cite{authors_vrachimis_leakdb_2018}.

As these scenarios are limited, simulating custom ones with a flexible choice of anomalies, e.g., leakages or sensor faults is a valuable option when one wants to evaluate a potential solution. The standard simulation tool is EPANET \cite{rossman_epanet_2000} which is commonly used through the python package WNTR \cite{klise_software_2017}. To simplify the process of generating multiple scenarios with different types of anomalies in easy-to-use workflows, we provide the Automation Toolbox for ML in water Networks (ATMN)\footnote{\url{https://pypi.org/project/atmn/}}. 
Next to the possibility of easily defining various scenarios containing multiple leakages and sensor faults, the toolbox provides parallelized, resource-aware automated generation with efficient storage of the simulation results. Besides, there is an easy-to-use data-loading API that is compatible with LeakDB, and ATMN  contains a visualization tool for \wdns including sensors, leakages, and sensor faults. Note that there also exist simulation tools for water quality simulations~\cite{kyriakou2023epyt}.

\textbf{Demands}
Next to the size and structure of the network, the demands influence how difficult the problem of leakage detection is in a given scenario. As stated in Section~\ref{sec:tasks} demand modeling itself is a difficult problem and thus an active research area. Many contributions evaluate their methodologies on artificial demands made up of simple sinusoidal curves or averages over a few collected exemplary demands.  One prominent example providing a large network and realistic demands is the BattLeDIM challenge \cite{vrachimis_battle_2022}.

\textbf{Evaluation}
Evaluating detection algorithms is usually done by considering the trade-off between the true positive rate and the false positive rate, which can be further analyzed by considering the ROC-AUC score. Since the goal is to keep potential water loss low, reporting detection delays~\cite{hu_review_2021} and water losses~\cite{vrachimis_battle_2022} is valuable.

\begin{landscape}
\begin{table}
    \caption{Summary of approaches for leakage detection and localization alongside their data requirements and a summary of how the stages of drift detection are realized.\label{tab:methods-overview}} 
    \centering
    \scriptsize
    \overviewtab{overviewtab.csv}{}{}
\end{table}
\end{landscape}
Regarding the localization of leakages, the goal is to pinpoint the location as precisely as possible. Measuring and comparing the topological distances is suitable~\cite{hu_review_2021}. Assuming that only scarce sensor data is available, analyzing whether the leakage lies in the area spanned by several sensors is a reasonable choice~\cite{vaquet2023localization} as then this area can be further investigated by traditional methodologies, e.g., sound-based localization techniques \cite{SITAROPOULOS2023101905}

\section{Approaches for Leakage Detection and Localization\label{sec:approaches}}

Considerable research has been conducted on leakage detection and localization~\cite{hu_review_2021}. In general, these approaches are frequently tested in very small benchmark networks and often rely on strongly simplified artificial demands. In this work, we will provide a categorization of approaches (i) that perform leakage detection and/or localization based on pressure, flow, and demand measurements and (ii) that are evaluated in larger networks with realistic demands or real-world measurements used as we assume better generalization and scalability in these cases. We summarize them in Table~\ref{tab:methods-overview}. Due to space constraints, we only include the six best-performing contributions from the BattLeDIM challenge \cite{vrachimis_battle_2022}.

As discussed before, leakages or anomalies more generally can be modeled as \emph{concept drift}, i.e., as distributional changes in the data generating distribution \cite{gama2014survey}.
While some works attempt leakage detection as end-to-end ML, many other approaches implement some kind of drift detection scheme.

In end-to-end ML, e.g., \cite{sun_leak_2019,zhou_integration_2019}, observations $x_t$ are directly mapped to the output \enquote{leak/no leak} given training samples. However, this approach is suffering from a strong class imbalance. In our literature review, we only found solutions that were evaluated on very small networks and thus excluded them from further analysis as this indicates that end-to-end classification on more realistic and complex networks is too challenging.

Drift detection schemes aim at deciding whether distributional changes occur given a stream of data \cite{hinder2023things}. There are two options: one can apply \emph{supervised drift detection}~\cite{gama2014survey}, where one analyses the model loss of some inference model as a proxy, or \emph{unsupervised drift detection}~\cite{hinder2023things}, where one analyses the data distribution directly. Summarizing the body of work on leakage detection, we found that all approaches can be categorized as drift detection schemes. In the following, we will provide a survey on this task structured according to these options. 
Afterward, we focus on the assignment of leakage localization. A summary of all methods is provided in Table \ref{tab:methods-overview}.

\subsection{Prediction-Residual-Based Approaches\label{sec:approaches-prediction-based}}
\begin{figure}
    \begin{subfigure}{0.5\textwidth}
         \centering
         \tiny
         \begin{tikzpicture}[node distance=1em]
        \node (sample) {$x_1, \dots, x_t$};
        \node[rectangle,draw] (prediction) [below = of sample] {predict ($h$) \ref{sec:appr-pred-pred}};
        \node (xhat) [below = of prediction] {$\hat{x}_{t+1}$};
        \node (resid) [right =  of xhat] {residual};
        \node (x) [right =0.3em of sample] {$x_{t+1}$};
        \node[rectangle,draw] (detect) [below = of resid] {detect \ref{sec:appr-pred-det}};
        \node (pred) [below = of detect] {leak/no leak};
       \draw [-stealth] (sample.south) -- (prediction.north);
       \draw [-stealth] (prediction.south) -- (xhat.north);
       \draw [-stealth] (xhat.east) -- (resid.west);
       \draw [-stealth] (x.south) -- (resid.north);
       \draw [-stealth] (resid.south) -- (detect.north);
       \draw [-stealth] (detect.south) -- (pred.north);
        \end{tikzpicture}
         \caption{\Predbased approaches}
         \label{fig:appr-pred}
    \end{subfigure}
    \begin{subfigure}{0.5\textwidth}
    \centering
         \tiny
         \begin{tikzpicture}[node distance=1em]
        \node[minimum width=4em, align=center] (sample) {$x_1, \dots, x_n$};
        \node (dots) [right = of sample] {$\dots$};
        \node[minimum width=4em, align=center] (sample2) [right = of dots] {$x_{t-n}, \dots x_{t+1}$};
        \node[rectangle,draw] (prediction) [below = of sample] {process \ref{sec:appr-obs-process}};
        \node[rectangle,draw] (prediction2) [below = of sample2] {process \ref{sec:appr-obs-process}};
        \node (resid) [below =5em of dots] {residual};
        \node[rectangle,draw] (detect) [below = of resid] {detect \ref{sec:appr-obs-det}};
        \node (pred) [below = of detect] {leak/no leak};
      \draw [-stealth] (sample.south) -- (prediction.north);
      \draw [-stealth] (sample2.south) -- (prediction2.north);
      \draw [-stealth] (prediction.south) -- (resid.west);
      \draw [-stealth] (prediction2.south) -- (resid.east);
    
      \draw [-stealth] (resid.south) -- (detect.north);
      \draw [-stealth] (detect.south) -- (pred.north);
        \end{tikzpicture}
         \caption{\Obsbased}
         \label{fig:appr-obs}
    \end{subfigure}
    \caption{Schematic of approaches used in the literature}
    \label{fig:enter-label}
\end{figure}
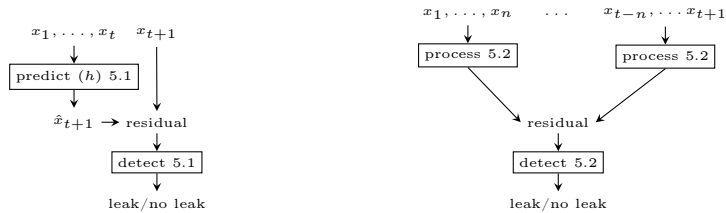

As visualized in Fig.~\ref{fig:appr-pred} supervised drift detection-based approaches, commonly also referred to as residual-based, fit a predictive model $h$ based on the normal operational state of the network and use it in the first step to compute the expected state of the network $\hat{x}_{t+1}$. In the second step, they analyze the residual of $\hat{x}_{t+1}$ and the measured -- possibly disturbed -- data $x_{t+1}$. Assuming $h$ fits the data this discrepancy is small. Under the assumption that $h$ does not generalize well to out-of-distribution samples, the model no longer fits the observed data if a leakage or another anomaly occurs, and thus the residual rises. From a theoretical viewpoint, these strategies are not generally suitable for monitoring tasks like anomaly detection since for ML models, the connection between model loss and drift is rather loose \cite{icpram}, i.e., a model would need to perfectly replicate the reality to not raise false alarms or miss leakages. This claim was empirically supported for leakage detection by \cite{vaquet2024investigating}.

\textbf{Prediction Phase\label{sec:appr-pred-pred}}
For the choice of the predicting model $h$ there are two main options: hydraulics-based approaches and \ml-based models\footnote{Note that in the water community, hydraulics-based approaches are called model-based and \ml-based approaches are called data-driven.}  \cite{hu_review_2021}.
\emph{Hydraulics-based} approaches realize $h$ as a hydraulic simulation model replicating the \wdn  \cite{li_fast_2022,steffelbauer_pressure-leak_2022,wang_multiple_2022,marzola_leakage_2022}.
This choice requires knowledge of the exact network topology, including pipe diameters and elevation levels. Further characteristics, e.g., pipe roughness coefficients, are usually calibrated using demand data. 
While building a precise hydraulic simulation model provides an accurate model of the network at hand, the requirements are very limiting: Usually, the precise network topology is not known, and real-time demands are not available. While these models can be easily adapted in case a new leakage has been detected and further analyzed, applying this strategy to new or evolving \wdns requires repeating the creation of the hydraulic simulation, a costly and time-intensive task. Thus, the generalizability of this approach is limited.

\emph{\ml} models constitute an alternative realization of $h$ and generally do not require knowledge of the precise network topology and demand measurements. However, usually, these methods assume the availability of leakage-free historical measurements of pressure (and flow) measurements.
\cite{daniel_sequential_2022} rely on linear regression to predict the pressure at each node. Similarly, \cite{laucelli_detecting_2016} apply evolutionary polynomial regression to keep an updated model of the network pressures and flows. \cite{romano_automated_2014} use a neural-network-based approach to predict the network state. Besides, \cite{vaquet2024investigating} apply some standard regression models to investigate the suitability of supervised drift detection. While the hydraulic-based methods take the demands as an input, this family of models generalizes over different possible network states which are caused by the demand patterns as inputs.

In contrast, there are some approaches leveraging the temporal dependencies in the data which are induced by the seasonality: \cite{wang_multiple_2022} propose to simulate future pressures and demands using time series analysis. They assume that the water consumption is composed of different components, i.e., trends, periodicities, and random factors, and use empirical mode decomposition (EMD) to decompose the data and consequently model normal leak-free operation. Similar to the hydraulic model-based approaches this method shares the weakness that it relies on demands for the EMD. \cite{jung_water_2015} are predicting group demands by applying Kalman filters. Those group demands can be used as input for the detection step. \cite{romero-ben_leak_2022} rely on a model predicting future demands.

Given knowledge of the topology and a less scarce sensor availability, one could explore ML models leveraging the spatial dependencies, for example, as done by \cite{ashraf2022gcn_wds} for missing value imputation. 

\textbf{Detection Phase\label{sec:appr-pred-det}}
The proposed methods for detecting the leakages range from visual inspection by a domain expert~\cite{marzola_leakage_2022}, simple thresholding \cite{romero-ben_leak_2022,laucelli_detecting_2016,romano_automated_2014}, over standard statistical approaches like CUSUM and Hotelling scores \cite{daniel_sequential_2022,jung_water_2015,wang_multiple_2022} to more complex approaches. \cite{steffelbauer_pressure-leak_2022} perform the CUSUM method on a dual model with artificial reservoirs at each node which captures the leaking water. An example of a more sophisticated approach using \ml is the contribution by \cite{li_fast_2022}. They perform a signal decomposition and apply $k$-means clustering on the trend component to find time periods containing leakages. 

\subsection{Oberservation-Residual-Based Approaches}
As shown in Fig.~\ref{fig:appr-obs}, observation-residual-based approaches that can be categorized as unsupervised drift detection analyze the differences of observation statistics \cite{hinder2023things}: the comparison is based on a compact description of the data characteristics observed in a reference and a current window. Anomalies are related to a significant change in these statistical compressions over time based on observed values. 

\textbf{Processing Phase\label{sec:appr-obs-process}}
In the first step, approaches aim to summarize and clean the data. \cite{xu_disturbance_2020} applies a wavelet transformation to obtain data smoothing. \cite{jung_improving_2015} eliminate demand patterns by applying z-scoring. \cite{loureiro_water_2016} discard large consumers and summarize the network flow data to estimate the total network consumption. \cite{vaquet2024investigating} leverages unsupervised drift detection schemes. To account for temporal dependencies in the data, they propose to consider two weekly windows to eliminate daily, weekly, and long-term patterns.

\textbf{Detection Phase\label{sec:appr-obs-det}}
The detection phase based on the residuals between the considered windows is mostly done by simple statistics: \cite{loureiro_water_2016} rely on robust statistics, e.g., the median absolute deviation. \cite{romano_automated_2014} rely on standard statistical methods and combine their findings with the prediction-based approach. \cite{jung_improving_2015} apply methods from statistical process control, e.g., Western Electric Company rules, CUSUM, exponentially weighted moving average control charts and Hotelling $T^2$ control charts.
\cite{xu_disturbance_2020} use an outlier detection by an isolation forest on incoming data and historical data samples which were collected at the same time of day (as similar patterns are to be expected).
\cite{vaquet2024investigating} evaluates a range of standard unsupervised drift detection schemes ranging from statistical tests, over virtual classifiers to block-based strategies.
This contribution has the advantage of not requiring historical leakage-free data. 

\subsection{Leakage Localization\label{sec:approaches-localization}}

Tackling the task of leakage localization there are mainly hydraulic-based approaches relying on the hydraulic simulators obtained in the first step of leakage detection as described in Section~\ref{sec:appr-pred-pred}. 
Here, the idea is to localize leakages by inverting the problem, i.e., by simulating leakages in different locations and minimizing the error over the possible leakage locations \cite{li_fast_2022,daniel_sequential_2022,marzola_leakage_2022,romero-ben_leak_2022}.
\cite{li_fast_2022} first narrow the possible leakage locations by a heuristic simulation relying on leakage flows. \cite{daniel_sequential_2022} uses linear programming to find the most realistic leakage location.
Additionally, \cite{li_fast_2022} rely on localizing one leakage at a time and then updating the hydraulic model with the leakage information.

However, there are also a few approaches considering this task through the lens of concept drift and analyzing their intermediate results closer. Some rely on an analysis of the residuals:
In the areas with sufficient sensor information, \cite{romero-ben_leak_2022} rely on graph-based interpolation and a geometric comparison of the measured and historical leak-free data to localize the leakage. Similarly, \cite{soldevila_data-driven_2019} rely on Kringing interpolation and Bayesian reasoning to increase the localization accuracy when compared to historical leakage-free data. 
\cite{steffelbauer_pressure-leak_2022} compute the residuals of the virtual flows described in the prediction section. Leakages are associated with a high Pearson correlation. \cite{vaquet2024investigating} analyze the results of feature-wise drift detection with a statistical test to find the sensors most affected by the leakage. Besides, \cite{vaquet2023localization} further analyze the concept drift by employing model-based explanations.

\subsection{The Role of Machine Learning in Leakage Detection and Localization}
So far \ml plays a limited role in solving leakage detection. For the briefly discussed end-to-end strategy, approaches suffer from too strong class imbalances. Looking at the more popular supervised approaches a mixed picture emerges. Considerable research has been conducted on using \ml models for predicting the network state under normal operation instead of relying on hydraulic modeling. As discussed before and can be seen in Table \ref{tab:methods-overview}, these methods are advantageous as they do not require real-time demands and the precise network topology. However, they rely on historical leakage-free data, and their detection capabilities are limited as shown theoretically in~\cite{icpram} and experimentally in \cite{vaquet2024investigating}. While still limited directly relying on unsupervised schemes seems very promising as recently shown by \cite{vaquet2024investigating}.

Regarding the localization, one heavily relies on hydraulic simulations which has the advantage that the entire network state is simulated, and thus, the precise leakage location can be determined. However, these strategies are not easily adaptable to changing and new networks and require real-time demands which is a strong limitation for their applicability in real-world applications. Relying on hydraulic measurements only, ML-based localization techniques seem a promising alternative assuming an increasing sensor availability.

We mainly focused on the key challenges raised in Section~\ref{sec:domain}. However, there are additional first works focusing on incorporating the additional constraints we discussed. For instance, there are first investigations on how to guarantee fairness in anomaly detection systems \cite{janine_fairness-enhancing_2023}. Besides, \cite{artelt2022explanation} investigate explainability in such systems. More research on these initial contributions needs to be conducted to develop holistic solutions incorporating all requirements. Additionally, more research on leveraging domain knowledge, e.g., by developing physical-informed ML approaches seems promising. 

\section{Conclusion\label{sec:concl}}

In this work, we presented open challenges in the domain of \wdns as an interesting and impactful endeavor for \ml research. In addition to presenting the main tasks and arguing why \ml can play an important role in the future, we discussed the particularities of the domain and proposed how to link them to the main tasks in WDNs. As a practical guide aiding first research efforts in the domain of \wdns, we presented suitable evaluation methods and provided an easy-to-use data generation tool. Besides, we considered leakage detection and localization as an exemplary task for which we provided a survey of the current methods.

\subsubsection*{Acknowledgment} 
We gratefully acknowledge funding from the European Research Council (ERC) under the ERC Synergy Grant Water-Futures (Grant agreement No. 951424). This research was supported by the Ministry of Culture and Science NRW (Germany) as part of the Lamarr Fellow Network. This publication reflects the views of the authors only.

\bibliographystyle{plain}
\footnotesize
\bibliography{water-bib}

\end{document}